\documentclass[runningheads]{llncs}
\usepackage[T1]{fontenc}
\usepackage{graphicx,verbatim}
\usepackage{amsmath,amssymb,amsfonts}
\usepackage{bm}
\usepackage{mathtools}
\usepackage{booktabs}
\usepackage[expansion=false]{microtype}
\usepackage{xcolor}
\usepackage{hyperref}
\usepackage{algorithm}
\usepackage{algpseudocode}
\usepackage{multirow}
\usepackage{subcaption}
\usepackage[table]{xcolor} 
\definecolor{myblue}{HTML}{D7E9FF}
\definecolor{mygreen}{HTML}{DFF5D8}

\usepackage{siunitx}
\begin{document}
\title{Few-Shot Concept Prompt Learning for Segmentation Foundation Models via Visual Grounding}
\titlerunning{Few-Shot Concept Prompt Learning}
%
\author{Rahul Venkataramani \and Rachana Sathish}

\authorrunning{F. Author et al.}
%
\institute{Advanced Technology Group, Bangalore}


\author{Rahul Venkataramani, Rachana Sathish}  
\authorrunning{Rahul Venkataramani et al.}
\institute{Advanced Technology Group, GE HealthCare \\
    \email{\{rahul.venkataramani,rachana.sathish\}@gehealthcare.com}}
  
\maketitle              
\begin{abstract}
Promptable segmentation foundation models (FMs) such as SAM3 and Medical SAM3 promise few-shot, interactively-specified segmentation for medical imaging through a natural language interface, yet their performance on clinical tasks falls well short of this promise. We posit that this shortfall is not an artefact of insufficient medical pretraining or imperfect prompt phrasing, but a structural limitation that will persist in any domain where paired image-text supervision is scarce, as it is across most clinical modalities. We further hypothesise that the limitation is specific to natural language as a control signal: a visually grounded prompt, learned directly from the target distribution, should recover the lost performance without additional image-text data or backbone retraining. We propose Few-Shot Concept Prompt Learning (FS-CPL), which learns a continuous concept prompt embedding $\mathbf{p}^* \in \mathbb{R}^{T \times d}$ from a small support set of $K$ image--mask pairs via mask supervision, with the encoder-decoder backbone frozen. Across four public benchmarks spanning ultrasound and endoscopy (BUSI, HC18, TN3K, CVC-Clinic), FS-CPL delivers absolute Dice improvements of up to $+0.62$ over canonical text prompts and is \emph{backbone-agnostic}: it lifts both vanilla SAM3 and the domain-specifically pretrained Medical SAM3, showing that visual concept prompting is complementary to in-domain pretraining. 

\end{abstract}

\section{Introduction}
\label{sec:intro}
\begin{figure}[ht]
    \centering
    \includegraphics[width=0.9\linewidth]{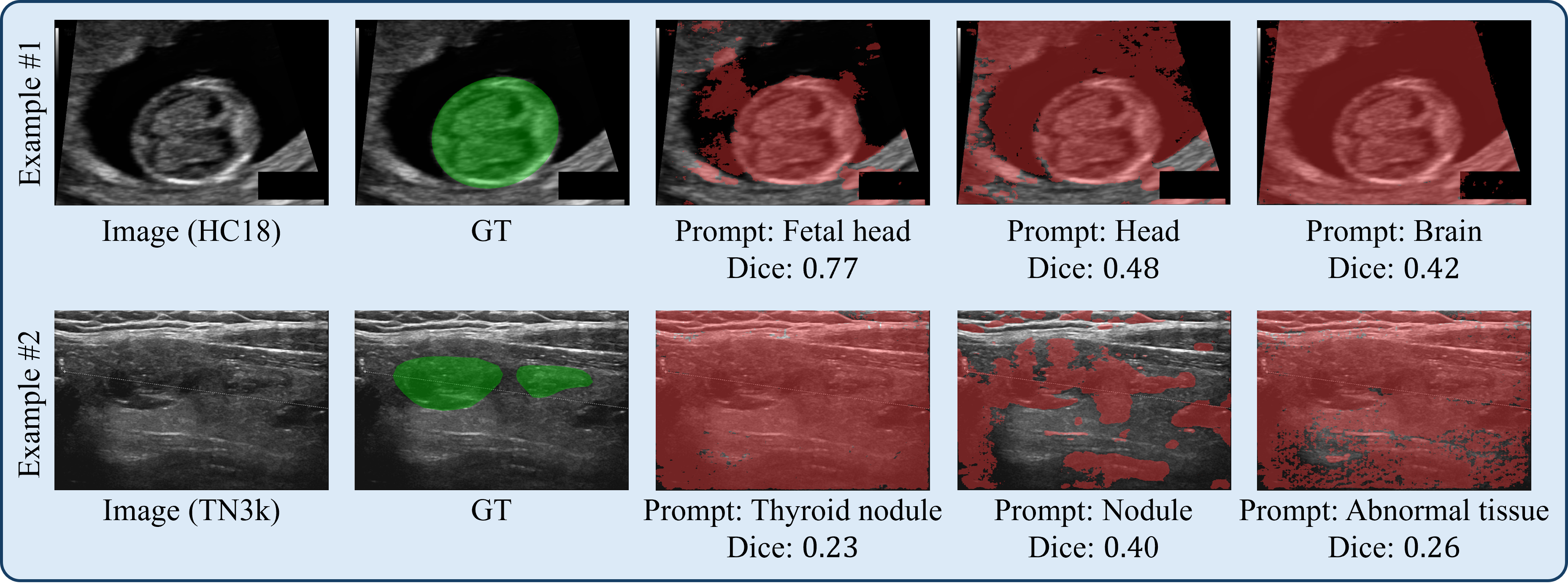}
    \caption{\textbf{Sensitivity of SAM3 to text prompts}. Example $\#1$ presents a sample image from HC18 dataset(fetal head ultrasound) and Example $\#2$ is of an image from TN3k dataset (thyroid ultrasound). The ground truth mask is overlaid in green over the input image and the predicted segmentation masks are overlaid in red.}
    \label{fig:intro}
\end{figure}

Text-prompted segmentation FMs (SAM3~\cite{Carion2025SAM3}, Medical SAM3~\cite{Jiang2026MedicalSAM3}) promise a leap in clinical utility by enabling segmentation on diverse tasks with limited language based supervision. Yet this promise is not practically realized in medical imaging through clinical language as medical terminology is heterogeneous, site-specific, and under-represented in tokenizers trained outside healthcare. Fig.~\ref{fig:intro} illustrates examples that demonstrates the brittleness of text prompts for ultrasound image segmentation. Prompts such as ``fetal head'', ``head'', ``brain'' produce different and sub-optimal masks for fetal head segmentation.

We introduce \textbf{Few-Shot Concept Prompt Learning (FS-CPL)} that adapts segmentation FMs to medical with minimal labeled data and eliminates the dependency on manual prompts. Inspired by text-prompted FMs, we \textit{prompt} FMs using a learned concept embedding. Instead of searching for the ``correct'' text phrase, we learn a concept prompt embedding ($\mathbf{p} \in \mathbb{R}^{T \times d}$, where $T$ is the number of prompt tokens and $d$ matches the backbone's text-embedding dimension) directly from a minimal support set of $K$ image-mask pairs. This approach effectively steers the frozen backbone via a learned embedding that captures the visual semantics of the target, creating a model that retains the agility of prompting without the fragility of syntax. We argue FS-CPL provides a powerful paradigm to prompt as the method optimizes over a larger manifold compared to text prompts, thereby accessing prompt manifolds unreachable by providing text inputs. 

Our evaluation across four diverse benchmarks spanning ultrasound and endoscopy (BUSI~\cite{al2020dataset}, HC18~\cite{van2018automated}, TN3K~\cite{gong2021multi}, CVC-Clinic~\cite{bernal2015wm}) demonstrates that FS-CPL systematically outperforms canonical text prompts, boosting Dice scores from $0.23 \rightarrow \mathbf{0.85}$ on CVC-Clinic and $0.78 \rightarrow \mathbf{0.91}$ on HC18 when applied to vanilla SAM3. Crucially, the same mechanism applied to the domain-specifically pretrained Medical SAM3 establishes, to our knowledge, a new state of the art among frozen-backbone methods on every dataset where Medical SAM3 is evaluable (TN3K, CVC-Clinic, HC18), demonstrating that FS-CPL is a \emph{backbone-agnostic} adaptation route that is complementary to in-domain pretraining rather than competing with it.

\subsection{Related Work}
\label{sec:related_work}

\noindent \textbf{Foundation Models in Medical Imaging.} 
SAM~\cite{Kirillov2023SegmentAnything}, SAM2~\cite{ravi2024sam2}, and medical adaptations like MedSAM~\cite{Ma2024MedSAM} segment distinct entities via geometric prompts (points, boxes), but remain interaction-dependent, precluding high-throughput automated analysis. To eliminate per-inference interaction, SAM3~\cite{Carion2025SAM3} and Medical SAM3~\cite{Jiang2026MedicalSAM3} introduced ``concept prompting'' via text, though clinical deployment is still hampered by ontology mismatch~\cite{li2026medical}. 

\noindent \textbf{Adapting Foundation Models with limited data.} Adapting FMs to medical images typically requires large annotated datasets. PerSAM~\cite{zhang2024personalize} overcomes this via a training-free, one-shot adaptation of SAM that uses a single reference image and mask to guide segmentation of the same object in new images. However, relying on a single exemplar makes the prior brittle to intra-class appearance variability.

\noindent \textbf{Prompt Learning vs. Prompt Engineering.} A parallel development in vision-language classification offers a solution to the lexical gap. \textbf{CoOp} (Context Optimization)~\cite{Zhou2022CoOp} and \textbf{CoCoOp}~\cite{Zhou2022CoCoOp} demonstrated that augmenting image embeddings with learnable continuous vectors significantly improves robustness to distribution shifts.

While CoOp focuses on image-level classification, we extend this philosophy to dense mask supervision. CoOp  optimizes textual content inside CLIP-like model to improve image-level recogntion. In contrast, we completely bypass the tokenizer, learn continuous token embedding from image-mask pairs, and feed them to a frozen decoder to drive pixel-wise outputs. Unlike PEFT methods that adapt the backbone, or AutoPrompt~\cite{Shin2020AutoPrompt} methods that search for discrete words, we learn a concept embedding that conditions a completely frozen segmentation backbone. This effectively bridges the gap between the automation of FMs and the precision of mask supervision, without the brittleness of natural language.

\section{Method}
\label{sec:method}

\begin{figure}[t]
    \centering
    \includegraphics[width=0.95\linewidth]{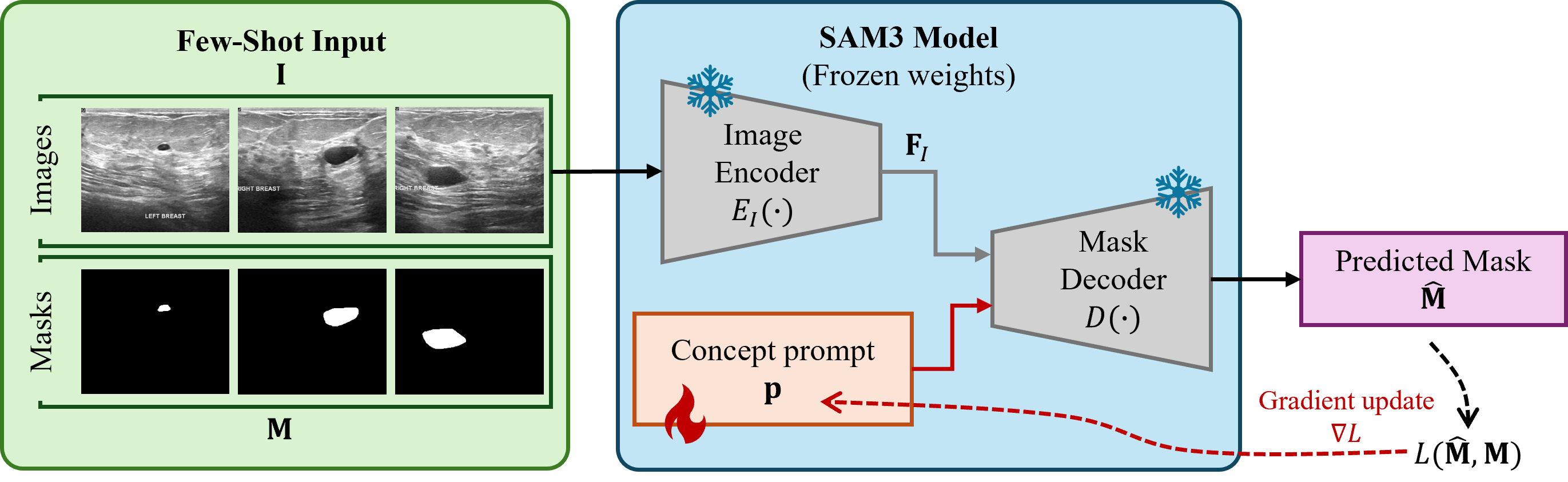}
    \caption{\textbf{Architecture of FS-CPL.} We bypass the discrete tokenizer by optimizing a continuous concept prompt $\mathbf{p}$ directly using mask supervision. The heavy-weight backbone (Image Encoder $E_I$, Decoder $D$) remains entirely frozen.}
    \label{fig:overview}
\end{figure}

We propose Few-Shot Concept Prompt Learning (FS-CPL), a technique to adapt generalist foundation models to medical concepts without parameter updates to the backbone. Our method, visualized in Fig.~\ref{fig:overview} formulates adaptation as an embedding optimization problem, seeking a continuous latent vector $\mathbf{p}$ that steers the frozen model to segment a target concept defined by a support set $\mathcal{S}$.\\

\noindent \textbf{Preliminaries.}
Let $\Phi = \{E_I, D\}$ denote a frozen promptable segmentation backbone, where $E_I(\cdot)$ is the image encoder and $D(\cdot)$ is a multimodal decoder. Given an input image $\mathbf{I} \in \mathbb{R}^{H \times W \times C}$, the encoder extracts visual features $\mathbf{F}_I = E_I(\mathbf{I})$. Standard prompting relies on a text encoder $E_T$ to map a discrete string $\mathbf{T}$ to an embedding $\mathbf{F}_T$. In our formulation, we discard $E_T$ and directly optimize a learnable soft prompt $\mathbf{p} \in \mathbb{R}^{T \times d}$, where $T$ is the token sequence length and $d$ is the embedding dimension. The segmentation logits are given by,
\begin{equation}
\mathbf{O} = D(\mathbf{F}_I, \mathbf{p}) \in \mathbb{R}^{H \times W}
\end{equation}
Here, the parameters of $E_I$ and $D$ are strictly frozen and only $\mathbf{p}$ carries gradients.

\noindent \textbf{Optimization Objective.}
We treat the concept prompt optimization as a few-shot learning task. A user provides a $K$-shot support set $\mathcal{S} = \{(\mathbf{I}_s, \mathbf{M}_s)\}_{s=1}^{K}$, containing representative image-mask pairs of the pathology. We then learn the optimal concept prompt $\mathbf{p}^*$ by minimizing the binary cross-entropy (BCE) loss over $\mathcal{S}$ as,
\begin{equation}
\label{eq:prompt_objective}
 \mathbf{p}^* = \operatorname*{arg\,min}_{\mathbf{p} \in \mathbb{R}^{T \times d}} \sum_{(\mathbf{I}_s, \mathbf{M}_s) \in \mathcal{S}} \mathcal{L}_{\text{BCE}}\Big( \sigma\big( D(E_I(\mathbf{I}_s), \mathbf{p}) \big), \mathbf{M}_s \Big)
\end{equation}
where $\sigma(\cdot)$ is the sigmoid activation. This process effectively leverages the mask decoder to find the optimal prompt embedding that produces the target mask. In our experiments, the prompt embedding $\mathbf{p}$ is optimized for $1000$ iterations ($N$) using Adam optimizer with a learning rate ($\eta$) of $0.001$.

\noindent \textbf{Inference.} 
At test time, the learned prompt $\mathbf{p}^*$ acts as a static concept key. For a query image $\mathbf{I}_q$, we simply compute the forward pass,
\begin{equation}
    \widehat{\mathbf{M}}_q = \mathbb{I}[\sigma(D(E_I(\mathbf{I}_q), \mathbf{p}^*)) \ge \tau]
\end{equation}
This approach is inductive as once $\mathbf{p}^*$ is learned, inference requires no further gradient updates or manual prompting, ensuring high throughput and reproducibility.

\section{Experiments}
\label{sec:experiments}

\subsection{Datasets}
To evaluate the robustness of FS-CPL across diverse imaging physics and anatomical scales, we employ four public benchmarks - (i) BUSI~\cite{al2020dataset} - breast ultrasound images with segmentation masks for lesion, (ii) HC18~\cite{van2018automated} - obstetric ultrasound images with  fetal head segmentations, (iii) TN3K~\cite{gong2021multi} - ultrasound images with segmentation masks for thyroid nodule and (iv) CVC-Clinic~\cite{bernal2015wm} - endoscopic images with segmentation masks for colon polyps. Unless an official split is provided, we generate disjoint patient-level train/test partitions.

\noindent\textbf{Pre-processing.} Images are first resized to the model’s required input dimensions of $1024 \times 1024$ pixels and then the intensity is scaled to [0,1] followed by normalization using mean and standard deviation of $0.5$.

\subsection{Few-shot protocol}
We adhere to a strict $K$-shot training protocol. Each experimental episode corresponds to a target concept (e.g., ``polyp'') and consists of two disjoint sets:
(i) a \textbf{Support Set} $\mathcal{S} = \{(\mathbf{I}_s, \mathbf{M}_s)\}_{s=1}^{K}$ and 
(ii) a \textbf{Query Set} $\mathcal{Q} = \{(\mathbf{I}_q,\mathbf{M}_q)\}_{q=1}^{Q}$. Given $\mathcal{S}$, FS-CPL optimizes the concept prompt embedding $\mathbf{p}$ for $N$ steps minimizing the segmentation loss (Eq.~\ref{eq:prompt_objective}). We utilize SAM3 and Medical SAM3 as the frozen backbones. The encoder and decoder weights remain frozen and adaptation is confined to the low-dimensional prompt embedding.

At inference, the learned prompt $\mathbf{p}^*\in \mathbb{R}^{T \times d}$ is applied to query images $\mathbf{I}_q$ without further test-time optimization. We set $T=32$ and $d=256$ in our experiments. Probability maps are thresholded at $\tau=0.5$ to yield binary masks $\widehat{\mathbf{M}}_q$. We analyze the data-efficiency of our method by varying $K \in \{1, 5, 10, 20\}$. No additional inference time is incurred due to our method over baselines.

\subsection{Baselines}

\textbf{Baselines.} We compare against (i) \emph{Canonical Text (Zero-Shot)}—frozen FM queried with the ontology term; (ii) \emph{Ensemble Text}—aggregated embeddings over template/synonym variations; (iii) \emph{Medical SAM3}~\cite{Jiang2026MedicalSAM3}, a medically-adapted FM; and (iv) \emph{PerSAM}~\cite{zhang2024personalize}, where a prototype from masked-mean-pooled FPN features over $K$ supports yields a top-10\% cosine-similarity bbox prompt for SAM3's grounding decoder.

\noindent\textbf{Evaluation metrics}: We report Dice similarity coefficient (Dice) and Intersection-over-Union (IoU) as primary metrics. All metrics are reported as mean $\pm$ 95\% CI half-width across the samples in the test split.


\section{Results and Discussion}

\begin{figure}[h]
  \centering
  \includegraphics[width=0.98\linewidth]{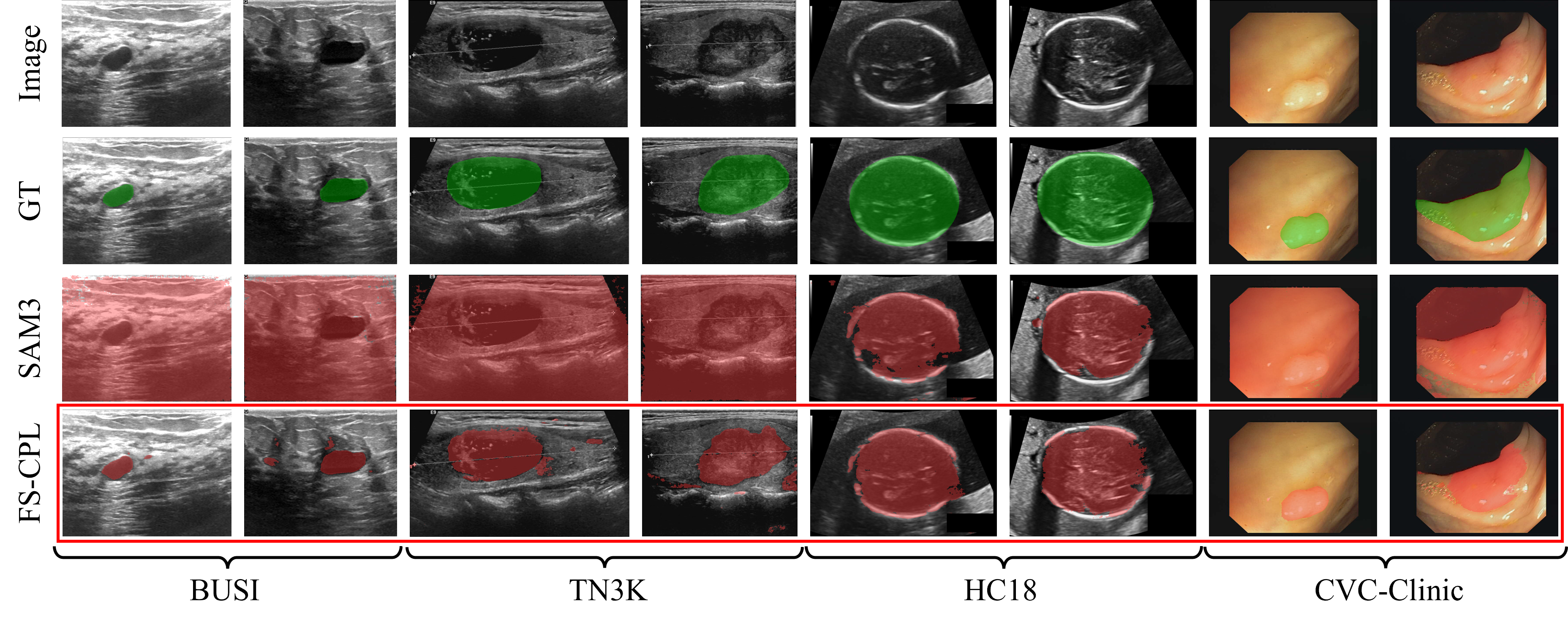}
  \caption{\textbf{Qualitative comparison}. Figure shows image, ground truth (GT), canonical text baseline (SAM3), FS-CPL result across four datasets. The GT mask is overlaid in green over the input image and the predicted segmentation masks are overlaid in red.}
  \label{fig:qualitative}
\end{figure}

Table~\ref{tab:main_results} presents quantitative evaluation on the test split of four datasets. 
Canonical prompts are: (i) BUSI - ``breast lesion'', (ii) HC18 - ``fetal head'', (iii) TN3K - ``thyroid nodule'', and (iv) CVC-Clinic - ``colon polyp''.  For ensembling, we use a fixed set of $10$ related prompts and average their text embeddings. Qualitative comparisons on sample images with ground truth (GT) are shown in Fig.~\ref{fig:qualitative}.

\begingroup
\setlength{\tabcolsep}{4pt}

\begin{table}[]
  \centering
  \caption{
    Few-shot segmentation performance for $K=20$. 
    Entries show mean $\pm$ 95\% CI half-width computed as $1.96\,s/\sqrt{n}$ where $s$ is the  std. dev. across test images and $n$ is the number of test images
(shown in parentheses).  Medical SAM3 is not included in the BUSI dataset evaluation, as this dataset was used during the model’s training.
    Best method per metric is highlighted in \colorbox{mygreen}{green} and second-best in \colorbox{myblue}{blue}.
  }
  \label{tab:main_results}

\begin{tabular}{l l l l}
    \toprule
    Dataset & Prompt Method & Dice $\uparrow$ & IoU $\uparrow$ \\
    \toprule

    \multirow{3}{*}{BUSI (437)}
      & Canonical & 0.13 $\pm$ 0.01 & 0.07 $\pm$ 0.01 \\
      & Ensemble  & 0.12 $\pm$ 0.01 & 0.07 $\pm$ 0.01 \\
      & PerSAM  & 0.33 $\pm$ 0.02 & 0.23 $\pm$ 0.02\\
      & \cellcolor{mygreen}FS-CPL (ours, SAM3)
        & \cellcolor{mygreen}0.50 $\pm$ 0.04
        & \cellcolor{mygreen}0.43 $\pm$ 0.03 \\
    \midrule

    \multirow{5}{*}{TN3K (614)}
      & Canonical & 0.20 $\pm$ 0.02 & 0.13 $\pm$ 0.02 \\
      & Ensemble  & 0.20 $\pm$ 0.01 & 0.13 $\pm$ 0.01 \\
      & PerSAM  & 0.24 $\pm$ 0.02 & 0.16 $\pm$ 0.03 \\
      & Medical SAM3
        & 0.41 $\pm$ 0.02
        & 0.33 $\pm$ 0.02 \\
      & \cellcolor{myblue}FS-CPL (ours, SAM3)
        & \cellcolor{myblue}0.59 $\pm$ 0.02
        & \cellcolor{myblue}0.48 $\pm$ 0.02 \\
      & \cellcolor{mygreen}FS-CPL (ours, Medical SAM3)
        & \cellcolor{mygreen}0.61 $\pm$ 0.02
        & \cellcolor{mygreen}0.49 $\pm$ 0.02 \\
    \midrule

    \multirow{5}{*}{CVC-Clinic (592)}
      & Canonical & 0.23 $\pm$ 0.01 & 0.14 $\pm$ 0.01 \\
      & Ensemble  & 0.23 $\pm$ 0.03 & 0.14 $\pm$ 0.01 \\
      & PerSAM  & 0.33 $\pm$ 0.02 & 0.23 $\pm$ 0.01 \\
      & \cellcolor{myblue}Medical SAM3
        & \cellcolor{myblue}0.88 $\pm$ 0.02
        & \cellcolor{myblue}0.81 $\pm$ 0.02 \\
      & FS-CPL (ours, SAM3)
        & 0.85 $\pm$ 0.02
        & 0.71 $\pm$ 0.02 \\
      & \cellcolor{mygreen}FS-CPL (ours, Medical SAM3)
        & \cellcolor{mygreen}0.91 $\pm$ 0.10
        & \cellcolor{mygreen}0.84 $\pm$ 0.10 \\
    \midrule

    \multirow{5}{*}{HC18 (800)}
      & Canonical & 0.78 $\pm$ 0.01 & 0.67 $\pm$ 0.04 \\
      & Ensemble  & 0.60 $\pm$ 0.02 & 0.46 $\pm$ 0.01 \\
      & PerSAM  & 0.48 $\pm$ 0.01 & 0.33 $\pm$ 0.01 \\
      & \cellcolor{myblue}Medical SAM3
        & \cellcolor{myblue}0.93 $\pm$ 0.01
        & \cellcolor{myblue}0.87 $\pm$ 0.01 \\
      & FS-CPL (ours, SAM3)
        & 0.91 $\pm$ 0.01
        & 0.84 $\pm$ 0.01 \\
      & \cellcolor{mygreen}FS-CPL (ours, Medical SAM3)
        & \cellcolor{mygreen}0.95 $\pm$ 0.07
        & \cellcolor{mygreen}0.92 $\pm$ 0.08 \\
    \bottomrule
\end{tabular}

\end{table}
\endgroup

FS-CPL consistently outperforms both text-prompting baselines across all datasets, with the largest margins on BUSI, TN3K, and CVC-Clinic. PerSAM also lacks FS-CPL by a big margin. We argue this gap is \emph{not} primarily attributable to vocabulary mismatch. Even when the target term is well represented in the BPE vocabulary\footnote{\url{https://github.com/facebookresearch/sam3/blob/main/sam3/assets/bpe\_simple\_vocab\_16e6.txt.gz}}, as is the case for ``head'', ``brain'', and ``skull'' on HC18, the canonical prompt still trails FS-CPL by a substantial margin (0.78 vs.\ 0.91 Dice). The bottleneck instead lies in the \emph{visual-task distribution} over which the FM has been supervised. Tokens such as ``head'' are predominantly associated, during pretraining, with visual instances drawn from natural-image domains, whose appearance statistics differ markedly from speckle-dominated, low-contrast ultrasound. Textual conditioning alone cannot bridge this distributional gap, irrespective of how faithfully the term is tokenized. Consistent with this account, prompt ensembling yields negligible improvement over a single canonical prompt: enlarging lexical coverage offers no remedy when the failure mode resides in the visual conditioning rather than in the wording. 

FS-CPL circumvents this limitation by grounding the prompt directly in the target visual distribution through mask supervision, obviating the requirement for a clean lexical handle on the concept. This benefit is, moreover, \emph{backbone-agnostic}. On vanilla SAM3, FS-CPL closes most of the gap to Medical SAM3 despite the latter's extensive in-domain pretraining; on Medical SAM3, FS-CPL further improves upon its already-strong text-prompted baseline. Visual concept prompting and domain-specific pretraining are therefore \emph{complementary}: learned visual prototypes recover information that neither a domain-adapted backbone nor a richer prompt vocabulary supplies in isolation. FS-CPL thus constitutes a lightweight, inference-time mechanism that consistently extracts additional signal from whichever foundation model it is paired with, while requiring only a small support set and no parameter updates. 

\subsection{Effect of support set size}
\begin{figure}
  \centering
  \includegraphics[width=0.9\linewidth]{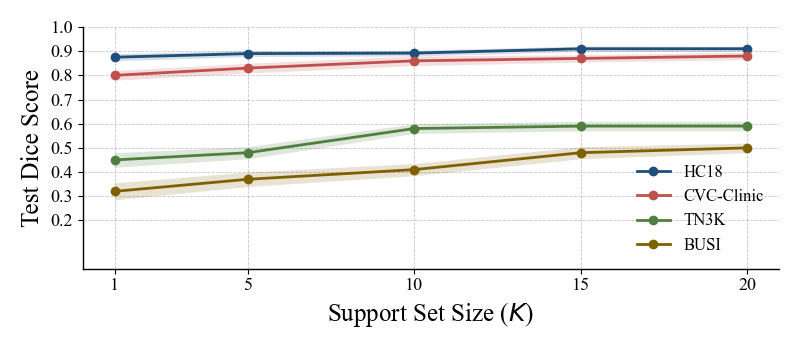}
  \caption{Effect of support set size ($K$) on performance. Figure shows the mean $\pm 95\%$ confidence interval of test Dice scores as a function of support set size ($K$).}
  \label{fig:support}
\end{figure}

Fig.~\ref{fig:support} analyzes the impact of the support set size $K$ on test Dice across the test datasets. We observe a clear trend of diminishing returns. The performance improves with increasing $K$ at small support sizes, but the gains become marginal beyond $K \approx 10$. This indicates that the proposed few-shot concept prompt embedding can be learned reliably from a small number of annotated image-mask pairs, and that additional annotations beyond $K=10$ provide limited benefit for these tasks. 

\subsection{Prompt embedding sensitivity and performance gains}
\label{prompt_sensitivity}
Text-encoder embeddings of synonymous prompts show only moderate mutual similarity, and the optimized concept embeddings $\mathbf{p}^*$ remain close to their text-based initialization yet yield substantially better segmentation; random initialization, in contrast, fails entirely. This indicates that $\mathbf{p}^*$ must lie within the text-embedding manifold, and that small, targeted displacements within it suffice to improve segmentation quality. For instance, ``colon polyp'' and ``colonic polyp'' have a cosine similarity of $0.69$, whereas the learned $\mathbf{p}^*$ initialized from ``colon polyp'' attains CS $\approx 0.98$ with the canonical embedding while delivering the gains reported in Table~\ref{tab:main_results}.

\section{Conclusion}
In this work, we introduced a few-shot prompt-learning approach for medical image segmentation with promptable FMs that learns a task-specific concept prompt embedding from a small support set of $K$ annotated image-mask pairs while keeping the encoder-decoder backbone frozen. Across three ultrasound datasets (breast lesion, fetal head, thyroid nodule) and one endoscopy dataset (colon polyp), the method consistently outperformed canonical ``best-guess'' prompts and prompt ensembles, showing that learning a compact concept representation is more effective than manually specifying or averaging text prompts.

We further showed that as few as $K=10$ annotated examples are sufficient to learn a concept prompt that generalizes well, with diminishing returns beyond this range on the studied tasks. A key advantage of the proposed method is that adaptation is achieved by optimizing only prompt parameters, reducing compute and memory demands and simplifying deployment and reproducibility. Future work will evaluate broader distribution shifts, multi-concept conditioning, and extensions to weak or noisy supervision.

%
%
%
\bibliographystyle{splncs04}
\bibliography{refs}
\end{document}